\documentclass{Interspeech}

\interspeechcameraready

\title{PhonemeFake: Redefining Deepfake Realism with Language-Driven Segmental Manipulation and Adaptive Bilevel Detection}

\author[affiliation={1}*]{Oguzhan}{Baser}
\author[affiliation={1}*]{A. Ege}{Tanriverdi}
\author[affiliation={1,2}]{Sriram}{Vishwanath}
\author[affiliation={1}]{Sandeep}{Chinchali}

\affiliation{Department of Electrical and Computer Engineering}{The University of Texas at Austin}{USA}
\affiliation{Department of Electrical and Computer Engineering}{Georgia Institute of Technology}{USA}
\email{oguzhanbaser@utexas.edu}

\usepackage{comment}

\begin{document}

\maketitle
{\renewcommand{\thefootnote}{*}\footnotetext{The first two authors have equal contribution.}}
\begin{abstract}
Deepfake (DF) attacks pose a growing threat as generative models become increasingly advanced. However, our study reveals that existing DF datasets fail to deceive human perception, unlike real DF attacks that influence public discourse. It highlights the need for more realistic DF attack vectors. We introduce PhonemeFake (PF), a DF attack that manipulates critical speech segments using language reasoning, significantly reducing human perception by up to 42\% and benchmark accuracies by up to 94\%. We release an easy-to-use PF dataset on HuggingFace and open-source bilevel DF segment detection model that adaptively prioritizes compute on manipulated regions. Our extensive experiments across three known DF datasets reveal that our detection model reduces EER by 91\% while achieving up to 90\% speed-up, with minimal compute overhead and precise localization beyond existing models as a scalable solution.
\end{abstract}

\section{Introduction}
Advancements in generative models have revolutionized the creation of synthetic content, enabling models to produce highly realistic audio, videos, and texts. Recent results, such as SORA's ability to generate high-fidelity photorealistic videos from text prompts, or GPT-like language models (LM) producing human-like conversations that are mostly indistinguishable from real interactions, highlight the capabilities of such technologies \cite{sora, brown2020language}. Although promising for applications such as creative media and virtual simulations \cite{arandcreativemedia}, it increases the risk of misuse. Sophisticated deepfake (DF) attacks now exploit subtle manipulations to deceive systems and humans alike, from segmental changes in audio to altered video frames \cite{had, mma}. Such developments expose critical vulnerabilities in existing detection mechanisms that often struggle to keep up with the growing complexity of generative models. Even beyond audio and video, missing subtle time-series forgeries (e.g., in EEG) has severe consequences \cite{eeg}. To effectively address these challenges, there is an urgent need for more comprehensive benchmarks and detection models that can reliably distinguish real from synthetic content, even in the face of highly refined DFs \cite{yan2023deepfakebench, baser2024securespectra}.

\begin{figure}[t]
    \centering
    \includegraphics[width=\linewidth]{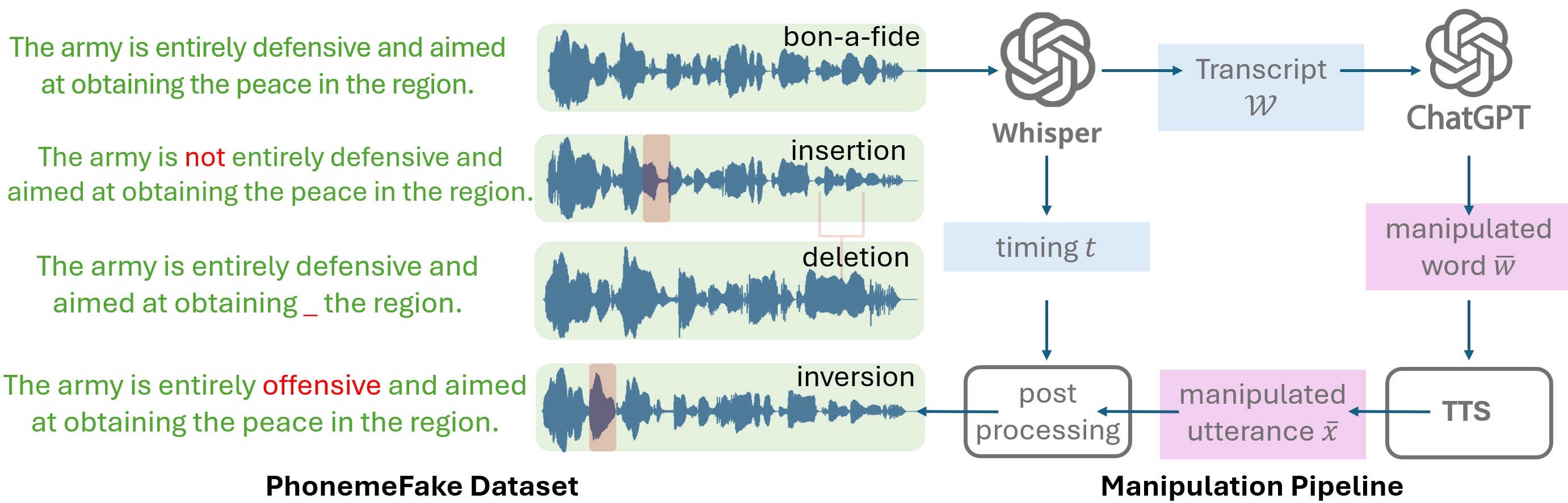}
    \caption{\small \textbf{What would be a more realistic DF attack and how does language reasoning help DF synthesis?} Our pipeline for generating segmentally manipulated DF audio samples integrates LM reasoning to emulate plausible and realistic DF attacks. Starting from a bon-a-fide audio recording, the transcription and timing are extracted using Whisper, and a specific target word or phrase is identified for manipulation with an LM. The manipulation process employs one of three strategies: \textit{inversion}, where a word is replaced with its antonym; \textit{insertion}, where a negating or altering phrase is added; and \textit{deletion}, where a critical word is omitted. These modifications are then synthesized using a TTS model to generate the manipulated utterance. The resulting DF retains the original flow, making detection challenging while reflecting real attack scenarios.}
    \label{fig:attacksynthesis}
\end{figure}

Despite significant advancements in DF detection, existing datasets and models fall short of addressing the complexities of real-world attacks. Current datasets predominantly feature synthetic content generated entirely from scratch, where the entire media (e.g., video, audio) is either completely original or entirely manipulated. While these datasets and the DF detection models trained with them achieve moderate success, they fail to replicate and detect the nuanced nature of real-world DF attacks. For example, a realistic attack may involve altering only the critical segments of a speech while leaving the surrounding content intact. This limitation is further evident in Fig. \ref{table:human} as it highlights that humans can reliably identify DFs in existing datasets but struggle significantly when faced with our dataset that introduces only segmental manipulations in critical utterances. Moreover, existing detection models are often biased towards predicting content as original when the initial segments appear authentic, reflecting a systemic lack of high-resolution detection capabilities for fine-grained manipulations. This bias arises from the training data, which disproportionately features media that is either fully original or fully synthesized. As a result, these models cannot effectively capture and localize subtle manipulations within a partially altered dataset, leading to lower accuracy in detecting realistic DF scenarios. Addressing these challenges requires a shift towards more representative datasets and detection models capable of handling localized, fine-grained manipulations, thereby bridging the gap between controlled research settings and real-world applications.

We \textbf{\textit{observe}} that the state-of-the-art (SoTA) DF detection models rely on datasets where the entirety of each sample is either fully synthetic or fully original \cite{baser2024securespectra,sasv2,scl}. This approach overlooks targeted real-world DF attacks, manipulating only the most crucial segments of a sample while leaving the rest intact. Consequently, existing models struggle to identify these fine-grained manipulations, as they are optimized to detect global inconsistencies rather than localized alterations. \textbf{\textit{Our key technical insight}} is in natural language and human communication, critical information is often concentrated in specific segments of a sentence or speech. For example, in a sentence, the words or phrases that carry high semantic or contextual weight are more likely to be the targets of DF manipulations. Based on this, \textbf{\textit{we hypothesize}} that prioritizing the analysis of these critical segments can improve detection efficiency and accuracy. This focus aligns with how real-world attackers are likely to craft manipulations, targeting the most impactful segments to maximize deception while minimizing detectable alterations. To address these concerns, we present a segmentally manipulated dataset and a search-then-infer detection model to target critical regions, enhancing DF detection accuracy and efficiency.

\noindent\underline{\textit{Literature Review}:} Existing audio DF detection methods rely on ML-based classification \cite{sasv2,scl,rev1,rev2,rev3,rawbmambda} or feature-level manipulations \cite{baser2024securespectra, water} (e.g., signatures, watermarking) but overlook semantic cues and language reasoning. While recent works use LM embeddings for robustness \cite{lang2, whisperfeatures, kale2024texshape}, they assume binary utterance-level classification, failing to detect segmental manipulations. PhonemeFake bridges this gap by explicitly targeting localized, semantically driven deepfake alterations. In the light of prior work, our contributions are three-fold:
\begin{itemize}
    \item \textbf{New Attack Vector}: We introduce a novel segmental DF attack vector that manipulates (e.g., inversion, deletion, and insertion) critical audio segments while preserving the surrounding content, better reflecting real-world attacks.
    \item \textbf{New DF Dataset:} We develop, PhonemeFake (PF), an easy-to-use, dataset\footnote{\url{https://huggingface.co/datasets/phonemefake/PhonemeFakeV2}} tailored for segmentally manipulated audio. This dataset significantly reduces the performance of the current SoTA detection models, highlighting their limitations.
    \item \textbf{Fast DF Detection Model}: We release a fine-grained detection method\footnote{\footnotesize\url{https://github.com/UTAustin-SwarmLab/PhonemeFake}}, PhonemeFakeDetect (PFD), optimized for segmental manipulations, achieving up to 90\% speedup over existing SoTA while reducing the error rate by up to 91\%.
\end{itemize}

\section{Method}
\begin{figure}
    \centering
    \includegraphics[width=0.85\linewidth]{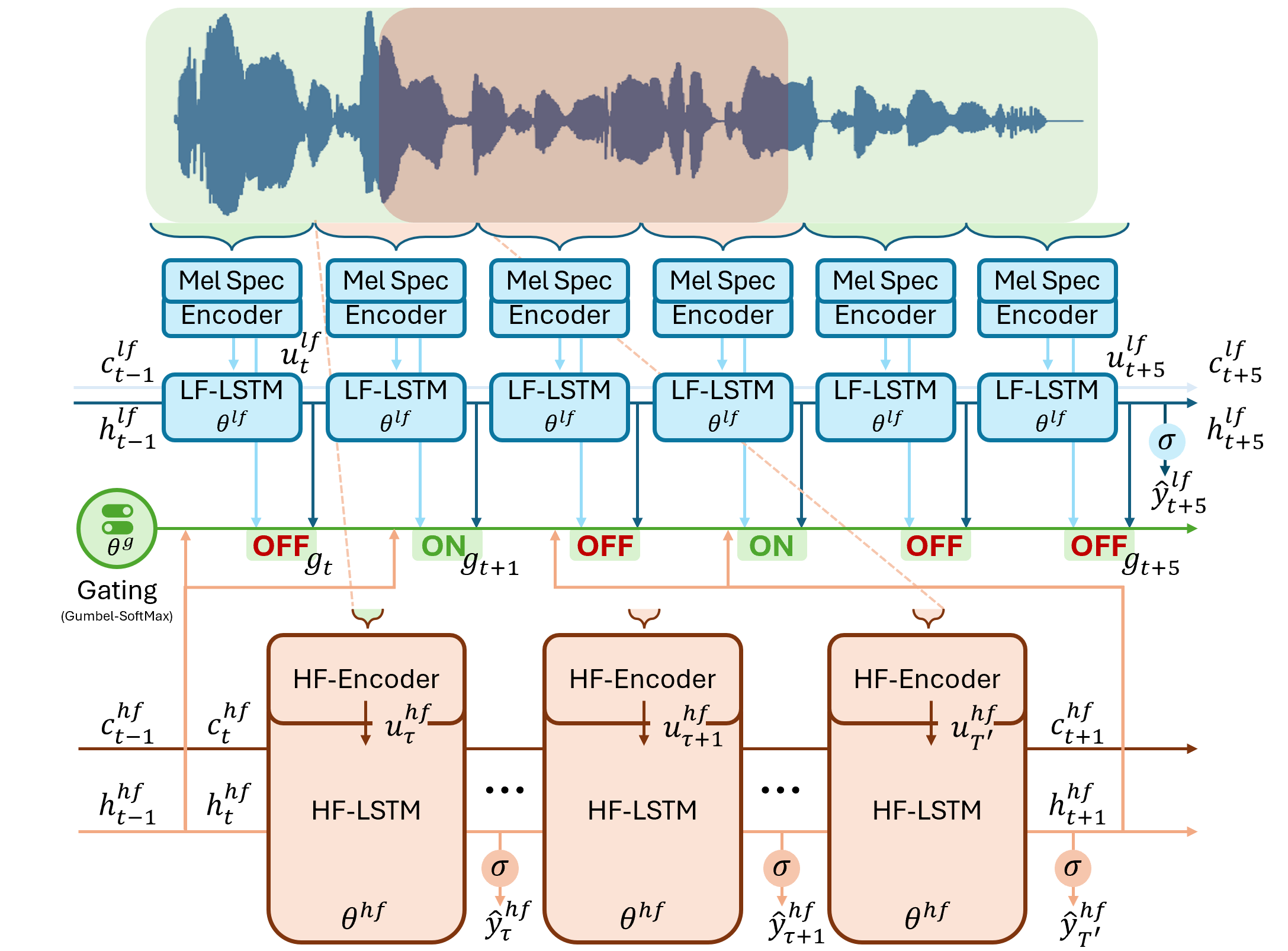}
    \caption{\small \textbf{How can we efficiently detect fine-grained DF manipulations?} Our bilevel detection model first uses an LF stream to identify RoI, then selectively activates an HF stream via Gumbel-Softmax gating for fine-grained analysis, ensuring high accuracy and resolution with minimal compute overhead.}
    \label{fig:detectionArchitecture}
\end{figure}

Our method comprises two key components: a generation pipeline that constructs a segmental DF audio dataset and a multi-level DF detection model designed to provide high-resolution timestamps for manipulation occurrences, as shown in Fig. \ref{fig:attacksynthesis} and Fig. \ref{fig:detectionArchitecture}, respectively.

\subsection{PFD: A Multi-Level Segmental DF Analysis Model}
Our method operates on two temporal resolutions, optimizing both computational efficiency and detection accuracy. The \textbf{Low-Frequency (LF) stream} processes the audio signal in large temporal windows, capturing broad contextual and temporal information. It serves as a preliminary lightweight filter, identifying regions of interest (ROIs) while maintaining low computational overhead. The \textbf{ high frequency (HF) stream} performs a detailed frame-by-frame analysis of the audio signal within the identified ROIs. Leveraging a more computationally intensive model, it extracts high-resolution features to scrutinize localized segments for subtle manipulations.This dual-stream approach balances global context awareness with localized precision, leveraging a dynamic gating mechanism to selectively activate the expensive HF stream only when necessary.

\begin{figure}
    \centering
    \includegraphics[width=0.7\linewidth]{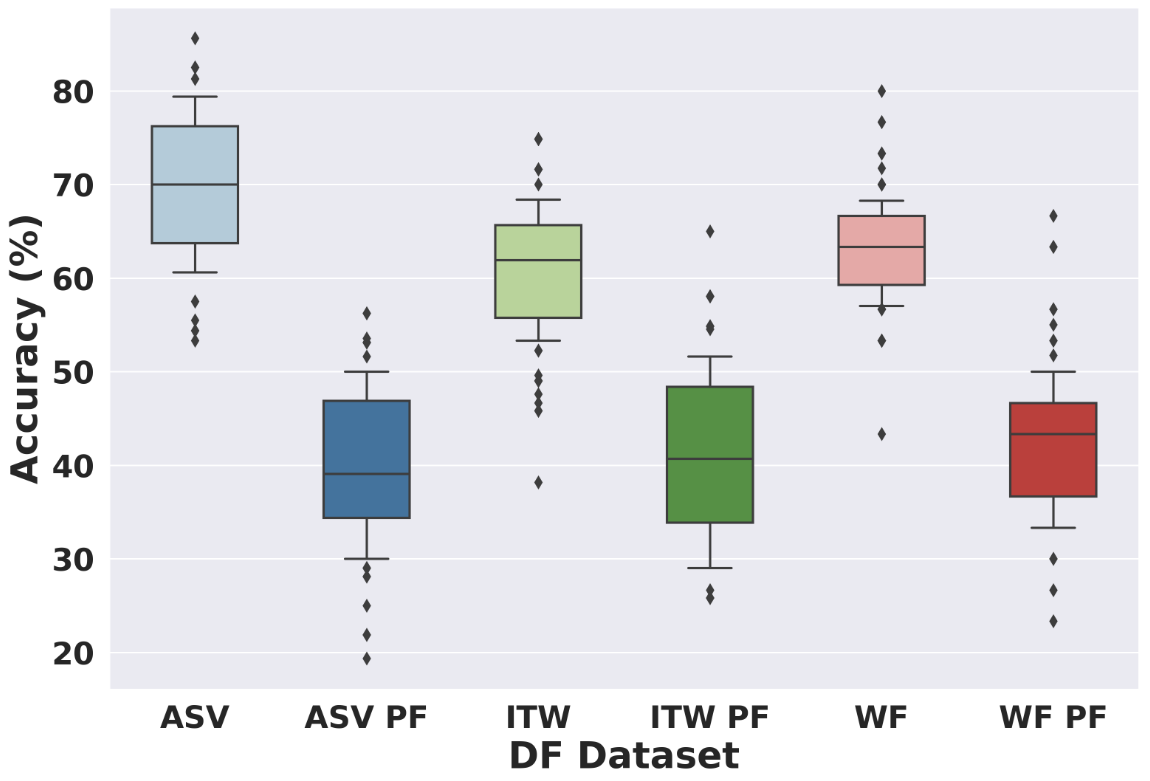}
\caption{\textbf{Do Current DF Datasets Deceive Humans?} Perception accuracy on 93 samples from 78 participants shows PhonemeFake is harder to detect than existing DF datasets.}

    \label{table:human}
\end{figure}

\subsubsection{LF Stream: Rough Analysis for ROIs}

The input audio signal is divided into \( T \) coarse temporal windows, each of size \( \Delta T \). For each window \( t \in \{1, \ldots, T\} \), the LF stream extracts low-dimensional features \( \mathbf{u}_t^{\mathrm{lf}} \in \mathbb{R}^{d_{\mathrm{lf}}} \) using a lightweight convolutional encoder. These features are then processed by an LF-LSTM network denoted as:
\begin{equation}
\mathbf{h}_t^{\mathrm{lf}}, \mathbf{c}_t^{\mathrm{lf}} = \mathrm{LSTM}^{(\mathrm{lf})}(\mathbf{u}_t^{\mathrm{lf}}, \mathbf{h}_{t-1}^{\mathrm{lf}}, \mathbf{c}_{t-1}^{\mathrm{lf}};\theta^{\mathrm{lf}}),
\end{equation}
where \( \mathbf{h}_t^{\mathrm{lf}} \) and \( \mathbf{c}_t^{\mathrm{lf}} \) represent the hidden and cell states of the LF LSTM at time \( t \). The prediction for LF frame $t$ is computed as $\hat{y}_t = \sigma(W^{\mathrm{lf}}\mathbf{h}_t^{\mathrm{lf}})$ where $W^{\mathrm{lf}}$ is a projection for LF predictions.

The LF-LSTM output serves as a preliminary filter, identifying ROIs where manipulations are likely. These ROIs are passed to the gating mechanism to decide whether the HF stream should be engaged for further analysis. 

\subsubsection{Gating: Dynamic Activation of the HF Stream}

The gating mechanism determines whether a detailed HF analysis is necessary for a given window \( t \). It takes the LF feature \( \mathbf{u}_t^{\mathrm{lf}} \), the LF-LSTM state \( \mathbf{h}_t^{\mathrm{lf}} \), the HF-LSTM state from the previous step \( \mathbf{h}_{t-1}^{\mathrm{hf}} \) and computes gating decision $\mathbf{b}_t$ as:
\begin{equation}
\mathbf{g}_t = \theta_g [\mathbf{u}_t^{\mathrm{lf}}, \mathbf{h}_{t-1}^{\mathrm{lf}}, \mathbf{h}_{t-1}^{\mathrm{hf}}],
\end{equation}
where \( \theta_g \) is a learnable projection matrix and \( [\cdot] \) denotes concatenation. $\mathbf{g}_t$ is a binary decision sampled from Bernoulli distribution parametrized by $\theta_g$, as shown by:
\begin{equation}
\mathbf{g}_t = 
\begin{cases} 
1 & \text{if HF analysis is required,} \\
0 & \text{otherwise.}
\end{cases}
\end{equation}
However, it is not differentiable through the argmax function. Using Gumbel-Softmax \cite{gumbelmax1, gumbelmax2} we approximate the discrete gating decision $\mathbf{g}_t$ to a differentiable random variable $G_t$. For binary prediction setting (\(K = 2\)), where class probabilities are proportional to $g_k$, sampling from the categorical distribution is $G_t = \text{argmax}_k \left(\log g_t^k + N_t^k\right)$ where $N^k = -\log(-\log(U^k))$ is Gumbel noise, and $U^k \sim \text{Uniform}(0, 1)$. This enables discrete sampling. However, argmax is nondifferentiable, making it unsuitable for backpropagation. To resolve it, the Gumbel-Softmax relaxation approximates the sampling with a differentiable softmax, shown as:
\[
G_t^i = \frac{\exp\left((\log g_t^i + N_t^i)/\mu\right)}{\sum_{j=1}^K \exp\left((\log g_t^j + N_t^j)/\mu\right)}, \quad i \in \{0, 1\}.
\]

\noindent The temperature $\mu>0$ controls the distribution sharpness. As $\mu \to 0$, the output approximates the \(\text{argmax}\)'s discreteness, while larger $\mu$ allows smoother and more exploratory decisions.

\subsubsection{HF Stream: Fine Resolution DF Detection}
When \( G_t^0 = 1 \), the HF Stream extracts high-dimensional features \( \mathbf{u}_\tau^{\mathrm{hf}} \in \mathbb{R}^{d_{\mathrm{hf}}} \) using a compute-intensive LSTM. Unlike LF-stream, HF-stream is processed in higher resolution steps $T'$ of size $\Delta T'=\frac{\Delta T}{T'}$. These fine features obtained from the narrow window $\tau$, concatenated with the LF features, are processed by the HF-LSTM to capture local temporal dynamics:
\begin{equation}
\mathbf{h}_{\tau}^{\prime\mathrm{hf}}, \mathbf{c}_{\tau}^{\prime\mathrm{hf}} = \mathrm{LSTM}^{(\mathrm{hf})}([\mathbf{u}_{\tau}^{\mathrm{hf}},\mathbf{u}_{t}^{\mathrm{lf}}], \mathbf{h}_{\tau-1}^{\mathrm{hf}}, \mathbf{c}_{\tau-1}^{\mathrm{hf}};\theta^{\mathrm{hf}}).
\end{equation}
If \( G_t^1 = 1 \), the old HF-LSTM state is spread without an update to keep the temporal context ($\tiny\mathbf{h}_{\tau}^{\mathrm{hf}}=\mathbf{h}_{\tau-1}^{\mathrm{hf}}, \mathbf{c}_{\tau}^{\mathrm{hf}}=\mathbf{c}_{\tau-1}^{\mathrm{hf}})$, as in:
\begin{equation}
    \mathbf{h}_{t}^{\mathrm{hf}} = \mathbf{G}_t
    \begin{bmatrix}
    \mathbf{h}_{T^\prime}^{\prime\mathrm{hf}}\\    
    \mathbf{h}_{t-1}^{\mathrm{hf}}
    \end{bmatrix},\mathbf{c}_{t}^{\mathrm{hf}} = \mathbf{G}_t
    \begin{bmatrix}
         \mathbf{c}_{T^\prime}^{\prime\mathrm{hf}}\\
         \mathbf{c}_{t-1}^{\mathrm{hf}}
     \end{bmatrix}
\end{equation}
We use \( \mathbf{h}_t^{\mathrm{hf}} \) to predict the high-resolution frame $\tau$ by $\hat{y}_\tau = \sigma(W^{\mathrm{hf}} \mathbf{h}_\tau^{\mathrm{hf}})$, where \( W^{\mathrm{hf}} \) is a learnable projection and \( \sigma(\cdot) \) is the sigmoid. The binary \( \hat{y}_t \) indicates whether the frame is real.

\subsubsection{Training Procedure}

The model is trained to optimize both detection accuracy and computational efficiency using a composite loss function $\mathcal{L}$. For each broad window, it sums the binary cross-entropy loss for both LF and HF classifications along with a fine-usage penalty scaled by mixture factor $\lambda$ to limit HF activations, given as:
\begin{equation}\footnotesize
\mathcal{L}_t = 
\begin{bmatrix}
G_t^0 & G_t^1 
\end{bmatrix}
\begin{bmatrix}
\lambda - \sum_{\tau=1}^{T'} \Big[y_\tau \ln(\hat{y}_\tau) + (1-y_\tau) \ln(1-\hat{y}_\tau)\Big]\\
-\big( y_t \ln(\hat{y}_t) + (1-y_t) \ln(1-\hat{y}_t) \big)
\end{bmatrix},
\end{equation}
 where \( y_t, y_\tau \) is the ground-truth label for frame \( t, \tau \) and \( \lambda \) controls the tradeoff between accuracy and efficiency. Then, the loss for each broad window is minimized as:
 \begin{equation}
\min_{ \{\theta^g, \theta^{\mathrm{lf}}, \theta^{\mathrm{hf}}, W^{\mathrm{lf}}, W^{\mathrm{hf}} \}} \,
\mathbb{E}_{\substack{g_t \sim \text{Bernoulli}(g_t; \theta_g) \\ (x, y) \sim D_{\text{train}}}}
\left[ \mathcal{L}_t
\right].
 \end{equation}
This optimizes detection, with the Bernoulli term adaptively triggering HF analysis only in suspected regions for efficiency. This bilevel strategy reduces the model's complexity from $O\left(T \cdot T' \cdot C_{\mathrm{HF}}\right)$ to $O\left(T \cdot C_{\mathrm{LF}} + p \cdot T \cdot T' \cdot C_{\mathrm{HF}}\right),$ where $C_{\mathrm{HF}}$ and $C_{\mathrm{LF}}$ denote the complexities of the HF and LF streams, respectively, and $p$ (tuned by $\lambda$) is the HF activation probability (approximately 0.1 in our experiments).\\


\noindent\textit{\underline{Why do we prefer LSTMs over transformers?}}\\
Our DF detection aims to localize subtle and segment-level manipulations. LSTMs inherently operate on temporal order and capture local dependencies, crucial for detecting fine-grained manipulations, whereas transformers compute global pairwise interactions, not necessary for our task, at a higher overhead. Also, ours worst-case complexity is $O(TT')$, whereas transformers incur $O((TT')^2)$ due to self-attention \cite{vaswani2017attention}. Lastly, LSTMs' inherent sequential processing naturally complements our bilevel framework's gating mechanism.

\subsection{PhonemeFake DF Generation}
\label{sec:synthesispip}
\noindent We design our synthesis pipeline to create realistic DFs leveraging LM reasoning, as described in Fig.~\ref{fig:attacksynthesis}. Let the input audio be $x(t)$, with its corresponding transcription $w = \{w_1, w_2, \dots, w_N\}$, where $w_i$ denotes the $i$-th word, and $t_i$ the timestamp. We manipulate the utterance $\bar{x}(t)$ by modifying the target segments of $x(t)$ followed by synthesis and fading. 

\noindent We extract audio $x(t)$'s transcription $w$ and word-level timing $t = \{t_1, t_2, \dots, t_N\}$ using the transcription model $\psi$ by $(w, t) = \psi(x;\theta_{\psi})$ \cite{whisper, baevski2020wav2vec}. Then, we identify the most semantically critical word ${w^* \in w}$ by prompting the LM $\Gamma$ as ${w^* = \Gamma(.,w;\theta_L)}$. Let $f_i(\cdot)$ denote the manipulation function operating in three modes $i\in\{1,2,3\}$: (1) Inversion: Replace $w^*$ with its antonym $w'$, yielding $f_1(w^*) = w'.$ (2) Insertion: Add a phrase $w^\dagger$ around $w^*$, such that $f_2(w^*) =  \{w^\dagger, w^*\}.$ (3) Deletion: Remove $w^*$, resulting in : $f_2(w^*) = \emptyset$. The manipulated transcription $\bar{w}$ becomes $\bar{w} = \{w \setminus w^*\} \cup f_{i}(w^*).$ Next, the TTS model $\phi$ synthesizes the manipulated segment $f_{i}(w^*)$ into audio $\bar{x}^* = \phi(x,f_{i}(w^*);\theta_\phi)$. To refine synthesized audio and ensure natural transitions, we apply a post-processing function $\mathcal{P}(\cdot)$, which handles tasks such as silence trimming and fade-in/out transitions in audio as $\bar{x}(t) = \mathcal{P}(x(t), \bar{x}^*, t)$. 

\noindent The resulting manipulated audio $\bar{x}(t)$ integrates subtle, localized manipulations while maintaining the natural prosody and flow of the original audio. This pipeline generates DFs that reflect semantically plausible and temporally coherent real-world attacks while remaining challenging for detection models.

\section{Experimental Setup}
Here, we outline the data, baselines, training, and test settings.\\
\noindent\textbf{\underline{Datasets}}: We evaluate our approach on three datasets. \textbf{WaveFake (WF)} \cite{wavefake} contains over 100,000 synthetic speech samples generated using six TTS models, offering a controlled benchmark focused on synthesis variations but it lacks real-world variability. \textbf{In-the-Wild Audio DF (ITW)} \cite{inthewild} captures DFs from real-world sources, introducing variations in recording conditions and background noise to assess the detection model generalization. \textbf{ASVspoof21 (ASV)} \cite{liu2023asvspoof}, part of the ASVspoof challenge, serves as a standardized benchmark for DF detection in speaker verification. \textcolor{black}{\textbf{SpoofCeleb (SC)} \cite{spoofceleb} features 800+ speakers and diverse TTS/VC attacks sourced from online media, offering ITW cases.} All consist of \textit{full DFs}, which are \textit{largely trivial for human perception}, as shown in Fig. \ref{table:human}. \textcolor{black}{Unlike those, \textbf{HAD} \cite{had} benchmarks partial DFs via manual tampering yet its forgeries sound robotic. We offer a language-driven pipeline for semantically coherent manipulations.}\\
\noindent\textbf{\underline{Detection Models}}: We benchmark our approach against three SoTA DF detection models. \textbf{SASV2} \cite{sasv2} learns a unified embedding space for speaker verification and spoof detection via \textit{multi-stage contrastive learning} and \textit{copy synthesis-augmented data}. \textbf{SCL} \cite{scl} improves generalization by treating \textit{re-synthesized DFs} as \textit{hard negatives} and applying \textit{balanced contrastive learning} for robustness against unseen attacks. \textcolor{black}{\textbf{AASIST} \cite{aasist} uses spectro-temporal graph attention on raw waveforms, achieving strong ASVspoof performance without handcrafted features.}
RawBMamba \textbf{RM} \cite{rawbmambda} leverages a \textit{bidirectional state-space model} to capture \textit{short- and long-range dependencies}, combining \textit{SincNet-based convolutions} with \textit{Mamba blocks} for feature extraction. However, all process full utterances globally, lacking the \textit{temporal resolution} to detect \textit{localized segmental manipulations}. While RM is the closest to our approach, it inefficiently processes the full input without selective refinement. In contrast, our \textit{bilevel model} adaptively focuses on the critical regions, achieving efficient and precise detection. \textcolor{black}{As no public models exist for audio forgery detection, we finetune a multi-modal forgery localizer, \textbf{MMA} \cite{ummaformer}, to audio with blank images for a fair comparison.}\\
\noindent\textbf{\underline{Hyperparameters}}: We compute PFD's LF stream over 1s windows for 64 mel bins and HF over 10ms for 128 mel bins, yielding a fine factor of 100. We model the LF and HF streams with 4-layer and 64-layer LSTMs (128 hidden size), respectively. The gating mechanism is regulated with a penalty factor $\lambda = 0.1$ to limit unnecessary HF activations, balancing accuracy and efficiency. We use Gumbel-Softmax with $\mu = 1$ for smooth gradient flow over gating. We employ Adam (lr=$10^{-5}$) with a scheduler, stopping after 20 epochs of no improvement in the validation loss. We run all on an 8x NVIDIA A100 cluster.\\
\noindent\textbf{\underline{Evaluation}}: All baselines were pretrained on ASV and fine-tuned on PF, ensuring equal exposure to our data. We retain originals as negatives to enforce subtle manipulation detection. We assess the methods using Equal Error Rate (\textbf{EER}) \cite{eer}. 

\section{Results}
Now, we analyze PF's impact on human and SoTA perception.

\begin{table}
    \centering
    \caption{DF Samples Synthesized from The Benchmarks}
    \begin{tabular}{c|cccc}
    \toprule
        Datasets & \textbf{WF} & \textbf{ITW} & \textbf{ASV21} & \textbf{PF} (Ours) \\
        \midrule
        Samples  & 11,201 & 15,333 & 16,361 & \textbf{42,895} \\
        Duration (h)  & 22.37 & 19.06 & 16.47 & \textbf{57.91} \\
    \bottomrule
    \end{tabular}
    \label{tab:dataset_comparison}
\end{table}

\noindent\textit{\underline{Do Existing DF Datasets Truly Deceive Human Perception?}}\\
\noindent To assess the perceptual detectability of DF speech, we conducted a user study with \textit{78 participants} spanning ages \textit{6 to 55}. Each participant evaluated around \textit{93 samples} of \textit{bon-a-fide, spoofed, and PF-variant} speech from benchmark datasets. The results in Fig. \ref{table:human} reveal that existing DF datasets, particularly ITW and ASV21, are easily distinguishable by human listeners, suggesting that they \textit{do not accurately represent realistic DF attacks capable of deceiving the public}. In contrast, our PF variants significantly reduced human detection accuracy consistently in the datasets by up to 42\%, while PF \textit{ critically altered the intended meaning}, demonstrating a much stronger similarity to real DF threats.\\
\noindent\textit{\underline{How Extensive and Realistic Is the PhonemeFake Dataset?}}\\
\noindent We generate DFs using the synthesis pipeline described in Sec.~\ref{sec:synthesispip}, leveraging bon-a-fide samples from the ASV21, ITW, and WF datasets. Our approach incorporates language-based manipulations, producing \textbf{57.91 hours} of synthesized DFs in \textbf{42,895 samples}. This process extends DF audio datasets such as ITW  by 273\% as shown in Table \ref{tab:dataset_comparison}, significantly increasing the diversity and realism of DF manipulations. Unlike fully synthesized DFs, our segmental approach maintains \textit{the original speaker's identity and prosody}, ensuring the manipulations remain \textit{subtle yet semantically impactful}. PF \textbf{87\%} likely attacks the top BERT-attended word in a sentence.\\
\noindent\textit{\underline{Can SoTA Detection Models Withstand Subtle Deceptive DFs?}}
Table \ref{tab:pf_comparison} presents the EER of SoTA DF detection models across the benchmark datasets and their segmental variations. We observe a \textit{consistent increase in EER} by up to 12.9 times when models are evaluated on our PF dataset, indicating their \textit{vulnerability to subtle yet semantically impactful manipulations} that retain the speaker's identity while altering the meaning. \\
\noindent\textit{\underline{How Does Our Bilevel Approach Fix This?}}\\  
\noindent Compared to SoTA models, our bilevel detection method demonstrates \textit{robust performance} with EER reduction of up to 91\%, with only an increase of up to 0.76 times in EER when transitioning from the original to the PF variants. This stems from our \textit{adaptive hierarchical processing}, where the LF stream identifies RoI, allowing the HF stream to focus compute on fine-grained manipulations. Prioritizing key segments over full utterances \textit{improves precision while maintaining efficiency}. This provides a \textit{scalable and reliable framework} for identifying \textit{realistic, segmentally manipulated DFs}. We observe that the HF stream is \textit{activated only 10\% of the time} during inference, proving the efficiency of the gating. Notably, the gate tends to activate precisely at the \textit{start and end of the manipulated segments, while remaining off in the middle}. Besides, LM attention is highly concentrated on altered segments, confirming that our method effectively captures the linguistic and acoustic inconsistencies introduced by DF modifications. Experimental extensions with SC, HAD, AASIST, and MMA are in the repo.

\begin{table}
    \centering
    \caption{Detection Performance in EER \% ($\downarrow$) across Datasets and Their PF Variants Denoting Segmenal Fakes.}
    \begin{tabular}{lcc|cc|cc}
        \toprule
         & \multicolumn{2}{c}{\textbf{ASV21}} & \multicolumn{2}{c}{\textbf{ITW}} & \multicolumn{2}{c}{\textbf{WF}}\\
        \textbf{Model} & Base &\textit{ PF} & Base & \textit{PF}& Base &\textit{PF}\\
        \midrule
        SASV2 & 27.67 & 54.57 & 17.91 & 55.11 & 8.18 & 10.19\\
        RM & 15.65 & 43.92 & 12.44 & 62.86& 33.14 & 94.96\\
        SCL & 2.63 & 21.37 & 4.51 & 22.79 & 1.79 & 23.11\\
        \midrule
        \textit{PFD} & 7.46 & \textbf{10.99} & 9.62 & \textbf{12.15} & 5.29 & \textbf{9.31}\\
        \bottomrule
    \end{tabular}
    \label{tab:pf_comparison}
\end{table}


\section{Conclusion}

We introduce a novel segmental DF attack vector, PhonemeFake, reducing human and ML detection accuracy by up to 94\%. Our human perception survey highlights that existing DFs are easily recognizable, whereas our manipulations produce more deceptive and realistic DF speech. To facilitate further research, we release a \textbf{comprehensive and user-friendly dataset} on Hugging Face, summing up to \textbf{57.91} hours. Also, we open-source a \textbf{computationally efficient bilevel detection framework}, capable of high-resolution timing prediction, achieving up to 90\% speed-up in inference while reducing detection EER by 91\% across multiple datasets. Future improvements include integrating LM attention into the gating to refine the DF timing, extending PF with more samples and cross-lingual attacks.

\section{Acknowledgments}
This work was supported in part by the National Science Foundation grants under No. 2148186 and No. 2133481. Any opinions, findings, and conclusions or recommendations expressed in this material are those of the authors and do not necessarily reflect the views of the National Science Foundation.



\bibliographystyle{IEEEtran}
\bibliography{mybib}

\end{document}